\documentclass[12pt,english]{paper}
\usepackage[T1]{fontenc}
\usepackage[latin9]{inputenc}
\usepackage{float}
\usepackage{amsthm}
\usepackage{amsmath}
\usepackage{graphicx}

\makeatletter

\newcommand{\lyxmathsym}[1]{\ifmmode\begingroup\def\b@ld{bold}
  \text{\ifx\math@version\b@ld\bfseries\fi#1}\endgroup\else#1\fi}

\providecommand{\tabularnewline}{\\}

\newcommand{\lyxaddress}[1]{
\par {\raggedright #1
\vspace{1.4em}
\noindent\par}
}

\makeatother

\usepackage{babel}
\begin{document}

\title{Emotional Responses in Artificial Agent-Based Systems: Reflexivity
and Adaptation in Artificial Life}

\author{Carlos Pedro Gonçalves}

\maketitle

\lyxaddress{Instituto Superior de Ciências Sociais e Políticas (ISCSP) - University
of Lisbon}

\lyxaddress{cgoncalves@iscsp.ulisboa.pt}
\begin{abstract}
The current work addresses a virtual environment with self-replicating
agents whose decisions are based on a form of \textquotedblleft{}somatic
computation\textquotedblright{} (soma - body) in which basic emotional
responses, taken in parallelism to actual living organisms, are introduced
as a way to provide the agents with greater reflexive abilities. The
work provides a contribution to the field of Artificial Intelligence
(AI) and Artificial Life (ALife) in connection to a neurobiology-based
cognitive framework for artificial systems and virtual environments\textquoteright{}
simulations. The performance of the agents capable of emotional responses
is compared with that of self-replicating automata, and the implications
of research on emotions and AI, in connection to both virtual agents
as well as robots, is addressed regarding possible future directions
and applications.\end{abstract}

\section{Introduction}

The combined research in Artificial Life (ALife) and Artificial Intelligence
(AI) allows one to address the coevolution of intelligence in artificial
systems, namely by providing artificial organisms with different degrees
of reflexiveness, one can test how basic toolkits of cognitive responses,
linked to reflexivity and adaptation to survival contexts, can be
more or less effective in terms of the adaptability of those artificial
organisms, and, therefore, provide better tools for simulation of
complex adaptive systems as well as solutions for technologies based
on evolutionary computation \cite{key-1,key-2}.

ALife, thus, allows one to deepen theoretical research in AI, introducing
a main central point: a basic survival problem common to living systems,
in the sense that life and death can be incorporated in the artificial
environment through birth (introduction of new artificial organisms),
living (the interactive and evolutionary dynamics of the artificial
organisms) and dying (the removal of artificial organisms). The possible
parallelism between artificial systems and the grounding work for
ALife can be traced to what can be called von Neumann's artificial
life conjecture, which is argued by von Neumann in his work on the
general and logical theory of automata \cite{key-3}.

Each artificial organism\textquoteright{}s AI, in turn, must be linked
to the management of the {}``life'' of the organism, anticipating
problems, avoiding dangers, adapting to environmental conditions and
trying to keep the organism alive. In this sense, we can deal with
artificial organisms as integrated systemic wholes that adapt to an
environment. A point that can be generalized to both artificial virtual
organisms as well as to physical agents like robots \cite{key-1}.

The AI of an artificial organism results from a system of programmed
rules, linked to the artificial organism\textquoteright{}s adaptive
interface with the environment and to that organism\textquoteright{}s
ability to: perceive, capture, select and process data and knowledge,
producing cognitive syntheses upon which the artificial organism can
act towards ensuring its survival. In this sense, the programming
of such an AI involves a form of basic somatic computation, a point
that is present in many ALife models \cite{key-1}.

One can speak of knowledge at the artificial agent's level, in the
sense of a resulting cognitive synthesis proceeding from an organizing
adaptive computational activity that is directed towards the agent's
survival and adaptation.

The programmed rules furnish a basic adaptive toolkit that must be
linked to a sensory system of the artificial organism, allowing it
to adapt to different situations. The entire structure of the organism\textquoteright{}s
makeup and AI introduce, at the artificial level, a basic dynamics
of intentionality in common with actual physical living systems, indeed,
dynamics of intentionality can be localized in any entity that may
be considered an agent dispositionally gifted of strategically operative
internal structures, a point that applies to ALife\textquoteright{}s
artificial organisms.

The programmed AI rules, being linked to an artificial organism\textquoteright{}s
survival, introduce, at the artificial level, a link between the \textquotedblleft{}life/living\textquotedblright{}
of the artificial organism and that organism\textquoteright{}s \textquotedblleft{}agency\textquotedblright{},
that is: in a similar way to an actual physical living organism, the
artificial organism\textquoteright{}s actions concur to a single purpose:
adaptation towards survival, such that the artificial organism is
directed towards the others, the environment and its own survival,
which resends to a notion of intentionality that can be invoked as
effectively applicable at the artificial level.

The present work addresses these issues, working with a variant of
Holland\textquoteright{}s Echo \cite{key-4,key-5,key-6}. We introduce
three types of self-replicating artificial organisms: 
\begin{itemize}
\item Plantoids: plant-like organisms that are fixed in a two-dimensional
game terrain that produce energy resources and that replicate, competing
for territory by spreading the replicas;
\item Cog-0 mobile agents: cog-0 are self-replicating mobile agents (or
m-agents for short) that feed from the plantoids and from other m-agents,
these cog-0 m-agents follow closely Holland\textquoteright{}s Echo
rules, they have a very basic level of agency, being mostly mechanical,
their adaptive interface results from their genetic makeup and their
interface structure;
\item Cog-1 mobile agents: cog-1 are also self-replicating m-agents, sharing
basic characteristics in common to cog-0, but are also programmed
with higher degrees of reflexivity, namely in terms of simulated emotional
responses, including: desire to feed, desire to replicate and fear. 
\end{itemize}
It is important to stress that the cog-1\textquoteright{}s emotional
responses are assumed by cognitive parallelism with basic human emotional
responses, that is, we programmed adaptive responses and levels of
perception that furnish cog-1 agents with an adaptive toolkit that
has a survival-prone calculatory basis, synthesizable by parallelism
in terms of what, in humans, are considered as basic emotional responses
in survival contexts.

The underlying conceptual ground for this work comes from von Neumann
\cite{key-3}, integrating Damásio's work in neurobiology, regarding
the role of emotions in human cognition and decision dynamics \cite{key-7}.
It is not, however, our purpose to simulate human agents in the cog-1
agents, instead, it is our purpose to introduce a basic adaptive computation
that leads to adaptive responsive patterns of cog-1s that have a similarity
to the adaptive responsive patterns associated, in humans, with emotional
responses in survival situations \cite{key-7}.

The reason to introduce such adaptive responsive patterns, in this
model, is to test the results of interactions and coevolution between
different artificial life forms with different types of AI, corresponding
to different degrees of awareness and adaptability, which may be relevant
for both simulation of artificial societies as well as for intelligent
systems\textquoteright{} design, regarding possible approaches to
introduce different degrees of autonomy and reflexivity that may help
in applications of adaptive computation, amplifying the adaptability
of intelligent systems%
\footnote{The issue of awareness and even self-awareness will be returned to
later on, when addressing cog-1s adaptive cognition.%
}.

In the model, for instance, while cog-0 automata \textquotedblleft{}simply\textquotedblright{}
feed, cog-1 agents are responsive to their internal body state, in
the sense that they evaluate their need to eat, synthesizable in the
calculation of a {}``desire to feed'', on the other hand, they simultaneously
evaluate the immediate surroundings and deliberate on what they should
do, so that their actions respond to basic survival questions: to
where should they move? Should they attack another replicator to steal
its resources, or should they avoid that attack, taking into account
the potential prey's strengths?

Motion for cog-1 agents is directed towards places with higher energy
resources (it always moves to \textquotedblleft{}greener pastures\textquotedblright{}).
In conflict situations, on the other hand, a cog-1 agent observes
a potential prey and evaluates whether or not attacking it is advantageous,
a calculatory dynamics is introduced, such that the cog-1 observes
the potential prey and anticipates the result of a confrontation,
signalized in terms of a fear versus desire cognitive dynamics: the
desire to feed from a potential prey overcomes the fear from physical
confrontation, whenever the prey is not stronger than the attacking
agent, on the other hand, if the potential prey is stronger, then
the fear of confrontation overcomes the desire to feed from the prey\textquoteright{}s
energy resources, and the cog-1 does not attack.

The simulated emotional responses are, in this case, linked to survival-prone
adaptive responses. The conceptual ground that allows us to address
these adaptive responses as emotional responses in artificial systems
is addressed in section 2. The purpose of the model is to evaluate
the emerging coevolutionary patterns between cog-0 and cog-1 agents
and under what conditions does cog-1 type of adaptive computation
become advantageous.

In section 2, the model is introduced. In section 3, the performance
of cog-1 versus cog-0 systems is addressed through model simulations
for different parameters. In section 4, the results of the model\textquoteright{}s
simulations and of the model's structure are addressed in a wider
context of the issue of simulating basic emotional responses in artificial
systems and the possible benefits of such simulation, to incorporate
in these systems greater degrees of autonomy and adaptiveness, so
that these systems can apply a basic toolkit to different situations,
increasing the \textquotedblleft{}plasticity\textquotedblright{} of
their adaptive computation. A discussion of the Echo model in connection
to the Kismet robot project is also provided in section 4.

\section{The model}

The model incorporates as basis Holland\textquoteright{}s Echo, so
that we have an artificial environment comprised of a terrain organized
in patches%
\footnote{We are using Netlogo's terminology, in this case, since further on,
the simulations analyzed will be from the Netlogo version of the model
available at http://modelingcommons.org/browse/one\_model/3950\#model\_tabs\_browse\_procedures
(the procedures that implement the model can be consulted at this
website).%
}, forming a square lattice with periodic conditions at the borders.
Each patch is occupied by a plantoid, which is characterized by a
reservoir capacity and energy resources generated by the plantoid
through its interaction with the local environment, each location\textquoteright{}s
plantoid has a location-specific (patch-related) energy reservoir
capacity. Plantoids play a role similar to the resource fountains
of Echo, however, they are self-replicating agents and each plantoid
possesses a characteristic chromosome chain, which determines the
plantoid\textquoteright{}s genetic identity.

Chromosomes are defined as per Echo\textquoteright{}s rules, in the
sense that they only share some of the characteristics of actual chromosomes.
While actual chromosomes have a much more complex relation with the
general structure of an organism, the chromosome in Echo\textquoteright{}s
artificial world is introduced with two fundamental characteristics:
the chromosome is the genetic material of the agent; the chromosome
determines the agent\textquoteright{}s interface and interaction patterns,
addressed in the model through tags that define these interactions,
the tags' contents match the segments of the chromosome\textquoteright{}s
chain \cite{key-5}.

In a full artificial environment's generality (that is, without trying
to match biology), one can assume a $n$ letter formal alphabet to
characterize the genetic code. In the simulations section we will,
however, consider Holland's four-letter code illustrative examples.
Since similar results were obtained for other values of $n$, we chose
the formal alphabet size that matches Holland's Echo.

Plantoids, being fixed in the terrain, compete for territory, so that
whenever, for a given plantoid, there is at least one neighbor, in
a Moore neighborhood%
\footnote{The Moore neighborhood structure is comprised of eight neighbors surrounding
the plantoid, in the two-dimensional square lattice with periodic
conditions at the borders.%
}, with less than half the resources of the plantoid, then, the plantoid
replicates and its replica replaces the neighboring plantoid. Replication
for plantoids takes place such that the new replica is born with the
same genetic code as the parent, with a (low) probability of a one
point mutation.

The presence of a new plantoid on a patch leads to a replenishment
of the energy resources at that patch to their maximum level, the
simulational assumption is that the new plantoid feeds from local
organic material leading to its energy resources being set at their
highest capacity possible for the patch where the new plantoid is
situated. The plantoids also have the capability to replenish their
energy resources at a certain replenishment rate so that at each computation
round, plantoids with energy below the local maximum capacity replenish
the resources at a fixed rate, which is one of the model\textquoteright{}s
parameters. Plantoids can lose energy due to interaction with cog-0
and cog-1 agents, which feed upon plantoids\textquoteright{} energy
resources.

Each m-agent (be it cog-0 or cog-1), as per Echo's framework \cite{key-4,key-5},
is born with a given chromosome chain that specifies two tags: an
offense tag and a defense tag, the m-agents also have an energy reservoir
capacity which is equal to a base maximum level plus an extra level
which is equal to the length of their genetic code, a point which
will be returned to further on. Besides the chromosome chain, cog-1
agents are also characterized by three more agent variables: desire
to feed, desire to replicate and fear, each of these variables are
updated along the simulation.

The setup of initial values for a simulation begins by the definition
of the initial territory and the plantoid configuration, the energy
resources' reservoir capacity of each plantoid depends upon the place
at which the plantoid is, so that this is a patch-related variable,
being linked to the interface between plantoids and the territory's
geographical characteristics.

In the simulations, the reservoir capacity is initially set randomly
for each patch between 1 and a maximum plantoids' energy reservoir
capacity (global parameter) with a discrete uniform distribution,
as stated, all plantoids are also set so that initially they all have
their energy resources at the highest reservoir capacity that the
patch allows.

Each plantoid\textquoteright{}s chromosome\textquoteright{}s length
is also set randomly with a discrete uniform distribution between
a length of 1 up to a maximum level which corresponds to one of the
model's global parameters (called, in the model's code, {}``max-gen'').
The initial values for the m-agents are set such that the chromosome
is built in two stages: the offense tag chromosome and the defense
tag chromosome. Both chromosomes\textquoteright{} lengths are set
randomly with uniform distribution between a length of 1 up to the
previously mentioned maximum level ({}``max-gen'').

In the general setting, as stated, the plantoids' and m-agents genetic
codes are defined as sequences taken from a \emph{n}-letter formal
alphabet, at each point in the chromosome sequence a letter is chosen
randomly with equal probability from the alphabet's letters so that
for a plantoid or a m-agent with a chromosome of length $m$, the
chromosome sequence will result from an equiprobable random draw from
the formal language of all $m$-length words.

For a m-agent, the full chromosome is a single chain comprised of
the offense tag sequence followed by the defense tag sequence which,
taken together in this way, make up the m-agent's chromosome sequence.

The reservoir capacity of the m-agents is set equal to a global value
of m-agents\textquoteright{} energy reservoir capacity (called, in
the model's code, \textquotedblleft{}agent-max-reservoir\textquotedblright{})
plus an additional capacity equal to the length of their genetic code,
this is linked to replication procedures, in the sense that replication
only takes place (with a some probability) if the m-agent's energy
resources exceeds a replication threshold, which is set, for each
m-agent, equal to the length of that agent's genetic code. The m-agents\textquoteright{}
reservoir is set initially to its full capacity.

In a simulation, an initial number of m-agents is set to an even number
divided in two halves, one half composed of cog-1 agents and another
half composed of cog-0 agents, this equal number of cog-1 and cog-0
agents is explained by the fact that given an equal number of cognitive
types we aim to evaluate which cognitive type tends to win the evolutionary
race and with what chances, depending upon relevant global parameters
set at the beginning of the simulation.

At the beginning of the simulation%
\footnote{This corresponds to an initial setup procedure.%
}, the cog-1 m-agents set their initial desire to replicate to 0 and
evaluate their internal state. The evaluation of internal state procedure
programs the cog-1 m-agents to be able to assess their internal energy
requirements, synthesized in an emotional response of desire to feed,
the desire to feed will be defined as an internal response of the
agent to the proportion of energy reservoir that is empty (called
proportion of energy needs, expressed by the one minus the quotient
between the cog-1\textquoteright{}s energy resources and the cog-1\textquoteright{}s
total reservoir capacity) the greater the proportion of energy needs
is, the greater is the desire to feed.

In the model, the cog-1s color changes (physiological response) depending
on the level of desire to feed, thus, the hungrier they are, the paler
the color is set, darker colors correspond to cog-1s that have less
desire to feed, this is programmed as physiological changes in their
artificial bodies as a consequence of the level of desire to feed.

One can speak, in this case, as argued next, of a {}``self-awareness''
associated with a calculus of survival, in the primitive sense of
awareness which resends to the notions of alertness and caution%
\footnote{This is the Old English common root with wary, with the sense of being
prudent, alert. The root can be traced to the Proto-Germanic {*}ga-waraz,
{*}ga (intensive prefix) and waraz (wary, cautious).%
}, primitively, awareness indicates a reinforced alertness, attention,
caution, systemically and cognitively, awareness can be linked to
an intentional directedness towards an object linked to objectives
of survival.

{}``Self-awareness'' can, therefore, be stated of any system\textquoteright{}s
cognitive dynamics in which the system addresses and regards itself
with an alertness that regards the system\textquoteright{}s own survival.
One can speak of {}``self-awareness'' with respect to artificial
systems, whenever they possess the capability of performing an internal
evaluation regarding the sustainability of their activity or any aspect
that is relevant for their continued functioning. This is not a {}``self-awareness''
of the same nature as human consciousness, nor intended to be a line
towards a simulation of human-like consciousness, rather, regarding
the cog-1 agents we are dealing with artificial intelligent systems
in an artificial environment with programmed cognitive responses that
furnish these agents with responses that are aimed at self-preservation,
including an ability to evaluate their internal state, their relation
to the environment and what they must do in order to survive in the
artificial {}``world''.

The cog-1 artificial agents, in this case, are programmed to evaluate
their internal energy resources and respond with a level of desire
to feed, the desire to feed allows for a linkage between the need
to replenish their resources and the future actions of feeding, thus,
the artificial cognitive system is equipped with an ability to evaluate
its internal state of energy needs and reflect this in its color and
in any future decisions regarding its need for feeding, which implies
the introduction of a basic level of self-awareness in the artificial
system.

Taking into account what is stated above about human-like {}``self-awareness''
and human-like emotions, along with the fundamental differences between
these and the nature of the artificial system under research, the
emotional responses can, still, be assumed and programmed inspired
by the human neurobiological examples, as stated before, through the
introduction of a type of somatic computation that plays a similar
role to the basic emotions and somatic markers identified in human
cognition regarding basic survival scenarios as researched by Damásio
\cite{key-7}, thus, in this case, the need to feed is synthesized
by a trigger alert for the agent, computed from its energy needs,
which warns the agent and plays a role in further adaptive cognitive
dynamics.

The simulation proceeds in rounds. At the beginning of each round
the plantoids replenish their energy resources. In this case, if the
maximum reservoir of a plantoid at a certain patch is not filled,
then, its energy reservoir is replenished by a fixed amount of replenishment
rate, which is, in this case, taken as a global parameter of the model,
otherwise, if the plantoid\textquoteright{}s maximum reservoir is
filled, then, its energy resources are kept at their peak level, so
that resource depletion can only come from m-agents' feeding from
the plantoids' energy resources.

The second step, in a round, is resource collection by m-agents, in
this case, the procedure depends upon a m-agent being a cog-0 or a
cog-1. Cog-0 agents limit themselves to collect energy resources from
the plantoid on the patch they are at, cog-1 agents are more elaborate
in their deliberation process. Before collecting energy resources
the cog-1 m-agents begin by evaluating their internal state, so that
their actions depend upon their desire to feed. There is a choice
facing the cog-1:
\begin{itemize}
\item It may stay on the patch where it is and feed from the plantoid there;
\item It may move towards another patch, moving towards the nearest neighboring
place that contains the plantoid with the highest energy resources.
\end{itemize}
Each time a m-agent moves to a new patch (taking one step) its energy
reservoir drops by one unit, thus, moving consumes energy, which means
that a cog-1 faces a decision that may cost it one unit of energy
and, in the limit, its life, since death, for a m-agent, takes place
whenever its energy reservoir reaches zero or drops below zero.

The desire to feed, for a cog-1 m-agent, signalizes to the cog-1 how
close it is to death, thus if $R_{c}(i)$ is the \emph{i-th} m-agent\textquoteright{}s
maximum reservoir capacity and $R(i)$ is the cog-1's current energy
level, then, if that m-agent is a cog-1 agent, the agent's desire
level $D(i)$ is given by the formula:
\begin{equation}
D(i)=1-\frac{R(i)}{R_{c}(i)}
\end{equation}
The desire to feed is updated every time the cog-1 evaluates its internal
state. Solving the above equation for $R(i)$ we get:
\begin{equation}
R(i)=\left(1-D(i)\right)R_{c}(i)
\end{equation}
the closer to $1$ the desire to feed is, the closer is the cog-1
to death. By programming the cog-1 with the ability to evaluate its
desire level and its maximum energy capacity $R_{c}(i)$, the cog-1's
emotional response to energy requirements warn it whether or not it
may move to a better location or stay on the location it is at. The
computation is the following: if $\left(1-D(i)\right)R_{c}(i)-1=0$,
then, the next step the cog-1 takes without feeding means its death,
therefore, the cog-1 does not take any step, on the other hand, if
$\left(1-D(i)\right)R_{c}(i)-1>0$, then, the cog-1 can take one step
to the patch that contains the plantoid with the highest amount of
energy resources%
\footnote{This corresponds to the moving to {}``greener pastures'' behavior.
The {}``greener pastures'' expression is literally so in Netlogo's
program procedures since there is an external color expression of
patches so that the greener the patch's color is set, the greater
is the amount of energy resources of the corresponding plantoid.%
}, after which the cog-1 evaluates again its internal state, which
triggers a new desire to feed level.

Risk, associated with death, is addressed by the cog-1 as an energy
exhaustion and the cog-1's programming allows for a systemic cognition
aimed at avoiding death. Moving to a {}``greener pasture'' or not
moving due to energy exhaustion are both linked to a systemic cognition
that is aimed at the organism's survival and permanence in the game.

In this way, we can state that the cog-1 artificial organisms are
knowledge producers, in the sense that any production of knowledge
is supported by an organizing activity that acts intentionally according
to the rules of the system, towards the sustainable survival of the
organism \cite{key-8}.

The laws of structure of an artificial organism are different from
a non-artificial one, artificial organisms are conditionned by the
specific systemic framework in which they are immersed, in this case,
an artificial environment, and by the program rules and restrictions
imposed upon the program, but by addressing each artificial organism
as a systemic whole with specific survival problems associated with
the structure of the artificial environment allows one to address
that organism's AI in terms of a cognition towards survival, which
approximates, in terms of a parallel correspondence, the intelligence
of the living organism and the intelligence of the artificial organism
in some fundamental aspects linked to survival, so that while the
artificial system has its own nature and rules that are specific to
it and that are distinct from the non-artificial system, it is still
possible to immerse it in a problematic context of survival and to
build a cognition towards survival that parallels that of the non-artificial
system, in terms of a functionality and adaptivity necessary to solve
survival and sustainability problems, this is what is done with the
cog-1 agents.

While cog-1 agents deliberate before feeding, the cog-0, as stated
before, just feed from plantoid that is on the patch where they are
located. The feeding itself follows the same procedure for both cog-0
and cog-1 agents, depending upon the interface of each m-agent in
relation to the plantoid and upon the number of m-agents feeding from
the same plantoid. The interface between m-agent and plantoid depends
upon the matching of their respective full chromosome chains, in particular,
upon the offense tag of the m-agent and the chromosome chain of the
plantoid, this is based upon Holland's Echo.

If the amount of resources of the plantoid at a given patch are greater
than zero, then, a comparison takes place between the offense tag
component of the chromosome of the m-agent and the plantoid\textquoteright{}s
chromosome. Thus, if the m-agent's offense tag is at least as large
as the plantoid\textquoteright{}s chromosome, then, assuming that
$k$ is the length of the plantoid\textquoteright{}s chromosome, the
first $k$ letters of the m-agent's offense tag are compared with
the chromosome string, if $n_{+}(i)$ is the number of matching letters
and $n_{-}(i)$ the number of non-matching letters, the matching level
for the \emph{i-th} m-agent is calculated by the difference $n_{+}(i)-n_{-}(i)$,
which is between $\lyxmathsym{\textendash}k$ and $k$. To obtain
a nonnegative interface score $s(i)$ we set the score as follows:
\begin{equation}
s(i)=0.5\cdot(n_{+}(i)-n_{-}(i)+k)+1
\end{equation}
which is between $1$ and $k+1$, the addition of one unit is explained
by the fact that, in feeding procedures, if all of the m-agents get
$0$ score we obtain a division by zero, which causes problems.

If the m-agent offense tag is shorter than the plantoid\textquoteright{}s
chromosome, then, if $k$ is the length of the m-agent's offense tag,
a penalty is included such that the score is halved: 
\begin{equation}
s(i)=\left[0.5\cdot\left(n_{+}(i)-n_{-}(i)+k\right)+1\right]\cdot0.5
\end{equation}
which is between $0.5$ and $(k+1)/2$.

The comparison between the chromosomes introduces an evaluation scheme
for interfaces between organisms, in this case, between the m-agent
and the plantoid, if the m-agent is a match to the plantoid and is
at least as complex as the plantoid in regards to chromosome length,
then, its interface is such that it gets the maximum score, it is
the best fit organism for feeding from that plantoid, however, if
the plantoid has a chromosome with a higher length than the m-agent,
even if the m-agent is a match it only matches a substring of the
plantoid's chromosome, which means that, in this case, it gets a penalty
for its lower length.

Other scoring schemes are possible, the scheme introduced is due to
the goal of introducing a selection scheme for chromosome length,
in the sense that m-agents with a longer offense tag length not only
have a greater diversity of possible plantoid food sources, they also
are able to get a differential advantage when feeding from the same
plantoid.

If the plantoid has energy resources, and a m-agent needs to feed
(which comes with an emotional response for a cog-1 and without an
emotional response for a cog-0), then, the m-agent feeds from the
plantoid's resources in a proportion that is equal to the agent's
score divided by the total score of the m-agents at that patch. Thus,
there is a selective pressure resulting from competition for access
to the plantoid.

In regards to feeding, the difference between the two types of agents
in terms of adaptive response is that the cog-1 will make a deliberation,
before feeding, whether it should move and where to, a deliberation
that depends upon the programmed emotional response synthesized by
the desire to feed. There is a greater reflexivity in the cog-1's
cognitive dynamics that translates in a greater adaptivity around
feeding. The second difference between the two cognitive types arises
in fighting.

After all m-agents have run through the feeding procedure, they enter
into a second set of procedures regarding predatory dynamics. All
m-agents whose energy reservoir is not full look for a prey to fight
with and from which they can steal energy resources. The cog-0 fight
with whatever other m-agent they find themselves facing, which means
that the cog-0 are fully exposed to the risk of either winning or
losing, in their predatory behavior.

In predatory mode, the cog-1 are more cautious, namely, the cog-1
look to the potential prey and anticipate the result of the confrontation
which means that the cog-1 run the scenario of the confrontation,
the result of this scenario triggers a dynamics of fear and desire,
such that the fear cognitive dynamics is, in this case, programmed
in terms of an emotional response to the confrontation itself, while
the desire cognitive dynamics results from the prospect of feeding
from the prey's energy resources.

The fear versus desire artificial cognitive processing, incorporated
in the cog-1's AI, allows the cog-1 agents to produce cognitive syntheses
aimed at adaptive responses regarding the anticipation of conflict,
signalizing to the cog-1 agent the risk involved in the conflict situation
which may allow the agent to avoid harmful conflicts and seek favorable
conflicts, reducing the exposure to conflict risk.

The first stage of conflict is as follows, a m-agent (be it cog-0
or cog-1), in predatory mode, seeks and meets with another m-agent
on the same patch. The conflict takes place in such a way that the
predator attacks, the prey defends and counterattacks, which means
that the predator also has to defend. Using Echo's rules, for the
transfer of resources from prey to predator (due to the predator's
attack), a comparison is made between the offense tag of the predator
and the defense tag of the prey, for the transfer of resources from
predator to prey (due to the prey's counterattack), a comparison is
made between the defense tag of the predator and the offense tag of
the prey.

The score is set in the same way as for the m-agent/plantoid interaction,
so that the scoring scheme above also holds for the matching between
an offense tag and a defense tag, instead of an offense tag and a
plantoid chromosome. Denoting the predator by the index $i$ and the
prey by the index $j$ we have the two scores, that we now explain:
\begin{equation}
s\left(ij\right)=\left[0.5\cdot\left(n_{+}\left(ij\right)-n_{-}\left(ij\right)+k_{ij}\right)+1\right]\cdot u_{ij}
\end{equation}
\begin{equation}
s\left(ji\right)=\left[0.5\cdot\left(n_{+}\left(ji\right)-n_{-}\left(ji\right)+k_{ji}\right)+1\right]\cdot u_{ji}
\end{equation}
In equation (5), $s\left(ij\right)$ is the score from the comparison
between the predator's offense tag and the prey's defense tag, $n_{+}\left(ij\right)-n_{-}\left(ij\right)$
is the matching level between the predator's offense tag and the prey's
defense tag and $k_{ij}$ is the length of the longest tag in a comparison
between the length of the predator's offense tag and the prey's defense
tag, $u_{ij}$ is a penalty which, in a similar way as equations (3)
and (4), is set to $1$ if the predator has the longest tag or if
the tag's have comparable size and to $0.5$ if the predator has the
shortest length tag%
\footnote{Selection pressure on the prey is, thus, for longer defense tags,
these present an advantage to the prey.%
}, in this way the scheme for m-agent/plantoid interaction is brought
now to the predator's attack to the prey.

The score $s\left(ji\right)$, in equation (6), is the score from
the comparison between the prey's offense tag and the predator's defense
tag, $n_{+}\left(ji\right)-n_{-}\left(ji\right)$ is the matching
level between the prey's offense tag and the predator's defense tag
and $k_{ji}$ is the length of the longest tag in a comparison between
the length of the prey's offense tag and the predator's defense tag,
$u_{ji}$ is a penalty which is set to $1$ if the prey has the longest
tag and to $0.5$ if the two tags are comparable in size or if the
prey has the shortest length tag%
\footnote{Selection on the prey is for longer offense tags used by a prey in
counter attack.%
}. A prey can have an advantage in the interaction if its score is
higher than the predator's, the transfer of resources due to the confrontation
is defined as follows:

\begin{equation}
T\left(ji\right)=R\left(j\right)\frac{s\left(ij\right)}{s\left(ij\right)+s\left(ji\right)}
\end{equation}
\begin{equation}
T\left(ij\right)=R\left(i\right)\frac{s\left(ji\right)}{s\left(ij\right)+s\left(ji\right)}
\end{equation}
In equation (7), $T\left(ji\right)$ represent the resources' amount
received by the predator, therefore, transferred from the prey to
the predator, the predator gets a proportion of the total resources
of the prey ($R\left(j\right)$) equal to its relative score in the
confrontation ($s\left(ij\right)/\left(s\left(ij\right)+s\left(ji\right)\right)$),
in the same way, from equation (8), it follows that the resources'
amount transferred from the predator to the prey, represented by $T(ij)$,
is such that the prey gets a proportion of the total resources of
the predator ($R\left(i\right)$) equal to its relative score in the
confrontation ($s\left(ji\right)/\left(s\left(ij\right)+s\left(ji\right)\right)$).
The amount of resources transferred depend upon the predator and the
prey's energy reservoirs level at the moment of confrontation and
upon their genetic makeup.

When conflict takes place which means the predator's new energy reservoir
level is set to:
\begin{equation}
\min\left[R\left(i\right)+T\left(ji\right)-T\left(ij\right),R_{c}\left(i\right)\right]
\end{equation}
where $R_{c}\left(i\right)$ is the predator's maximum energy reservoir
capacity, the formula $R\left(i\right)+T\left(ji\right)-T\left(ij\right)$
means that the predator receives energy resources from the prey, but
also loses energy resources to the prey, if that transference equals
or surpasses the predator's reservoir, then, the reservoir is filled
to its maximum capacity, on the other hand, if $R\left(i\right)+T\left(ji\right)-T\left(ij\right)\leq0$,
that is, if the energy resource loss surpasses both the resources
received and the agent's reservoir, then, the predator dies. For a
prey (cog-0 or cog-1) we get a similar result for the new energy reservoir
level:
\begin{equation}
\min\left[R\left(j\right)+T\left(ij\right)-T\left(ji\right),R_{c}\left(j\right)\right]
\end{equation}

Unlike cog-0 in predatory mode, who engage the prey no matter what,
cog-1 predators evaluate, first. the potential prey and anticipate
the result of the resource transfer, which means that they can form
an expectation as to the outcome of the confrontation. Two emotional
responses are involved here, the fear of confrontation is triggered
by the antecipatory perception of the outcome of the confrontation,
it is set, in this case, to the relative proportion of resources transferred,
so that the fear level is set as:
\begin{equation}
F\left(ij\right)=\frac{T\left(ij\right)}{T\left(ij\right)+T\left(ji\right)}
\end{equation}
thus, the cog-1 predator fears the confrontation by a level equal
to the relative proportion of resources transferred to the prey over
the total amount of resources transferred in the conflict. The cog-1
predator's desire, in turn, is set to:
\begin{equation}
D\left(ij\right)=1-F\left(ij\right)=\frac{T\left(ji\right)}{T\left(ij\right)+T\left(ji\right)}
\end{equation}
so that the predator fears the transference of energy resources to
the prey and desires the transference of resources from the prey.
If the fear surpasses the desire $F\left(ij\right)>D\left(ij\right)$,
then the cog-1 predator knows that the prey has the advantage in the
conflict, since it will transfer more resources to the prey than those
that it will receive from the prey, on the other hand, if the fear
does not surpass the desire $F\left(ij\right)\leq D\left(ij\right)$,
then the cog-1 predator knows that it has the advantage in the conflict,
since it will receive at least the same resources from the prey as
those that it will lose to the prey. If the fear of conflict is greater
than the desire, the cog-1 color turns yellow (artificial physiological
response) and the cog-1 avoids the conflict, if the fear of conflict
does not surpass the desire, the cog-1 color turns red (artificial
physiological response) and the cog-1 attacks its prey.

Cog-1 preys, under attack by another m-agent in predatory mode, also
make an analogous evaluation, calculating the fear level as:

\begin{equation}
F\left(ji\right)=\frac{T\left(ji\right)}{T\left(ij\right)+T\left(ij\right)}
\end{equation}
and the desire level as:
\begin{equation}
D\left(ji\right)=1-F\left(ji\right)=\frac{T\left(ij\right)}{T\left(ij\right)+T\left(ij\right)}
\end{equation}
the physiological response to fear versus desire is the same, even
though if attacked by the predator, the prey must defend itself%
\footnote{Modifications of the model might include the cog-1 prey's possibility
of escaping the predator with a certain probability, or depending
upon certain sections of the chromosome.%
}.

After a m-agent runs through its predatory stage it implements the
replication procedure. If the agent has gathered enough energy resources
surpassing the replication threshold (defined as the length of its
genetic code), then the agent replicates, with some probability%
\footnote{We do not assume that replication necessarily takes place. Thus, we
introduce a random component to replication, so that a m-agent may
or may not replicate. The replication probability is usually set to
a high value.%
}.

When replication takes place, it is such that the new m-agent receives
the same chromosome as the parent with a low probability of a one-point
mutation. This means that the offense and defense tags will, generally,
coincide, except in the cases of a one-point mutation. The new m-agent
also inherits the replication threshold, maximum reservoir capacity
and cognitive type of the parent with no transition probability from
cog-1 to cog-0 or from cog-0 to cog-1%
\footnote{This is done, in order to be able to compare the performance of the
two populations (cog-0 and cog-1). An evolutionary game where the
cognitive type could be changed would introduce another type of framework,
which can, however, be used in future simulations. In the present
case, the intention was to separate the genetic profile from the cognitive
profile in order to evaluate the differential advantage of different
AI systems, that is, the purpose was to evaluate whether, given the
same global conditions regarding environment and genetic profiles,
under which cases the cog-1 AI provides an advantage over cog-0, in
terms of adaptive performance.%
}. The parent gives half its energy resources to the newborn.

A number of physiological changes are introduced to cog-1 agents,
when they are replicating, namely, when their energy reservoir surpasses
their replication threshold this triggers their desire to replicate
(which is set to 1), and they change their color to pink (physiological
response). After replication their desire to replicate is reset to
0, and they reevaluate their internal state. Cog-1 newborns also evaluate
their internal state as soon as they are born.

After confrontation and replication, those m-agents whose energy resources
become depleted (agent reservoir of zero or below) die. The last procedure
left for m-agents is a moving procedure, but before moving, a new
procedure enters, here, into play, which is the plantoids' reaction
to m-agents' by releasing a poison that depletes the energy of the
m-agents by one unit, thus, m-agents that have lost all their energy
resources or m-agents that could not gather more than 1 unit of energy
resources will die after the plantoids poison release.

When a m-agent survives the whole sequence of energy resources harvesting,
predator/prey dynamics, replication and the poison release by plantoids,
then, that agent has survived the round so far, in this case, the
agent can move to another location%
\footnote{This is included, in the Netlogo model, in the poison release code
section, after the poison release procedure, in this sense, the moving
procedure by m-agents can also be considered an adaptive response
to the plantoids' poison release.%
}. Cog-1, as stated before, only move if they have enough energy to
do so, that is, if the energy is greater than 1 unit, since otherwise,
moving would lead to energy depletion, if they have enough energy
to move, as already stated, they move to the nearest patch with the
greatest energy resources. Cog-0 move no matter what and do it randomly,
which means that those cog-0 that have only one unit of energy will
die in the process of moving.

The last procedure of the model is named competition and is a plantoid
procedure, defining plantoids' replication rules. Thus, as stated
before, whenever a plantoid has at least one neighbor (in a Moore
neighboorhood) with less than half its energy resources the plantoid
replicates, replacing with its replica one of the randomly chosen
neighbors that conforms to this condition. Replication takes place
in such a way that the new plantoid has the same chromosome as the
parent but with a low probability of a one-point mutation. The new
plantoid replicator has an energy reservoir at the full capacity allowed
by the patch, which is justified by its consumption of organic materials
available at the patch so that the new organism is able to replenish
the energy resources to full capacity. 

These are the model's main procedures. We now analyze Netlogo's simulation
results.

\section{Simulations}

The model's coevolutionary dynamics is such that the m-agents play
a role in plantoid coevolution, by feeding off plantoids depleting
their energy resources they allow for plantoids to replicate due to
local energy resources' differences, which means that the m-agents
play a role in the plantoids' competition dynamics. On the other hand,
the plantoids also play a role in the m-agents coevolution, both at
the level of the genetically-determined interface (connected to feeding
procedures) as well as at the level of the killing of m-agents due
to poison release at the end of the round, before replication, which
means that only the replicators that are most efficient in their energy
gathering are able to survive.

A cog-1 m-agent has an evolutionary advantage over a cog-0 at the
moving, feeding and fighting levels. In moving, the cog-1 conserves
more energy since it only moves if it has enough energy do do so and,
when it moves, it goes towards {}``greener pastures'', which increases
its chances of gathering more energy resources during feeding. Although
this advantage does not always occur. Indeed, the {}``moving to greener
pastures'' procedure is a heuristic one, the cog-1 identifies plantoids
nearby, with greater amount of energy resources, however it is only
able to discover if its interface is fit when it starts to feed. This
is a purposeful assumption, since it allows for imprecise adaptation,
so that the cog-1 addresses its environment in terms of what it can
perceive, with insufficient information being present.

The {}``moving to greener pastures'' procedure also leaves room
for experimental expansion in terms of the cog-1's AI, namely higher
reflexivity levels going up the scale from basic emotional responses
to higher degrees of awareness with the formation of internal models
of the world, including the ability to form an internal map of the
territory built from the ability for geographic recollection. Since
the purpose of the current model is only to address basic emotional
responses we chose not to address, at this early stage, these higher
reflexivity levels, which will have to deal with expanded levels of
self-awareness%
\footnote{The agent has to be programmed to remember itself on that place before.%
}.

In conflict procedures, the cog-1 in predatory mode also has advantage
over the cog-0, since it only attacks if it anticipates an advantage
to do so, the cog-0 does not have this antecipatory ability, it sees
only the present. The reduced reflexive ability of the cog-0, that
is unable to reflect how close it is to death, meaning that it moves
even if that kills it and fights, even if that depletes more its energy
resources, make the cog-0 a very basic replicator, its selective advantage
is solely at the level of its offense and defense tags, and immediate
surroundings.

The cog-1, being always aware of how close it is to death, of its
feeding needs and of its surroundings and fighting advantages, is
able to gather and conserve more energy and so make it less probable
to be killed from poison release and other threats. Also a cog-1 with
a smaller genetic advantage in terms of offense and defense tags,
may be able to compensate through its more expanded adaptive cognitive
abilities than the cog-0.

Taking into account these factors it is expected that cog-1s, in general,
tend to dominate, in particular in the cases of harsh environments.
This is shown to be the case in repeated simulations: the cog-0 agents
tend to become extinct in harsh environments, while the cog-1 agents
can prosper, attaining sustainable population numbers.

Thus, for instance, in no simulation of the model have we found cog-0
surviving when the plantoids' maximum reservoir capacity is set to
10 units%
\footnote{Which means that each plantoid-harboring patch is initially set with
discrete uniform probability a maximum energy reservoir capacity of
up to 10 units.%
}, which is a small value that introduces a strong selective pressure
for m-agents who must survive in an environment of low abundance.
In this case, we always observed cog-0 becoming extinct, with cog-1
winning the evolutionary race over the cog-0. However, the resilience
of cog-0 populations strongly depends upon the world size and the
maximum chromosome length.

The world size tends to provide cog-0 agents greater chances of dispersing
and finding food sources, lowering the environmental stress. Another
factor of survival found in the simulation of the model is that cog-1
agents tend to spontaneously aggregate forming feeding colonies or
even nomadic groups that move in tandem (discussed further on). When
these groups are abundant they can compensate the environment's harsher
conditions and survive, cog-0 can find high energy resources in cog-1
colonies, however, these colonies also present a risk for the cog-0
in the sense that it can be fed upon by cog-1 agents in predatory
mode that only attack when they have the advantage.

The spontaneous aggregation of cog-1 agents in feeding colonies is,
in part, explained by the way in which they decide whether to move
or whether they should stay on one place, they only move if they find
an advantage, and they always move to the local place with the highest
energy resources, which means that they tend to aggregate on places
with higher energy resources and stay at those places or, alternatively,
move to a better place in the vicinities, but this motion is directed,
only dispersing in some instances where there are multiple local plantoids
with large amounts of resources. In this way, the cog-1 tend to aggregate
in groups that tend to move in tandem or in large feeding colonies.

Considering the Netlogo's base system of coordinates, which places
the origin at the central lattice site, and a square lattice of $-r$
and $+r$ extremes both horizontally and vertically, means that, for
different values of $r$, the lattice will be composed of $\left(2r+1\right)\times\left(2r+1\right)$
lattice sites (called patches in Netlogo). Table 1. below shows the
statistics for the time to extinction of cog-0 agents for 10,000 repeated
simulations with $r=2,3,...,10$, and plantoids' maximum reservoir
capacity of 10 units. While, in each group of 10,000 repeated simulations,
for each lattice size, the cog-0 agents always became extinct, the
mean time to extinction increases as the lattice size increases. Even
though the growth in mean seems to slow down. The median shows a more
stable behavior increasing for lattice sizes of 7x7 up to 13x13 and
then stabilizing from 13x13 up to 19x19, increasing again for the
21x21 lattice.

\begin{table}[H]
\begin{centering}
\begin{tabular}{|c|c|c|c|c|c|c|}
\hline 
{\small Lattice} & {\small Mean} & {\small 95\% Conf. Int.} & {\small Median} & {\small Skewness} & {\small Minimum} & {\small Maximum}\tabularnewline
\hline 
\hline 
{\small 7x7} & {\small 13.85} & {\small {]}13.74,13.97{[}} & {\small 13} & {\small 1.68} & {\small 4} & {\small 74}\tabularnewline
\hline 
{\small 9x9} & {\small 15.01} & {\small {]}14.89,15.12{[}} & {\small 14} & {\small 1.32} & {\small 3} & {\small 57}\tabularnewline
\hline 
{\small 11x11} & {\small 15.43} & {\small {]}15.32,15.55{[}} & {\small 14} & {\small 1.21} & {\small 3} & {\small 58}\tabularnewline
\hline 
{\small 13x13} & {\small 15.84} & {\small {]}15.72,15.95{[}} & {\small 15} & {\small 1.32} & {\small 3} & {\small 70}\tabularnewline
\hline 
{\small 15x15} & {\small 16.12} & {\small {]}16.01,16.24{[}} & {\small 15} & {\small 1.34} & {\small 4} & {\small 89}\tabularnewline
\hline 
{\small 17x17} & {\small 16.4} & {\small {]}16.28,16.52} & {\small 15} & {\small 1.19} & {\small 2} & {\small 69}\tabularnewline
\hline 
{\small 19x19} & {\small 16.44} & {\small {]}16.32,16.56{[}} & {\small 15} & {\small 1.16} & {\small 4} & {\small 60}\tabularnewline
\hline 
{\small 21x21} & {\small 16.66} & {\small {]}16.54,16.78{[}} & {\small 16} & {\small 1.13} & {\small 3} & {\small 64}\tabularnewline
\hline 
\end{tabular}
\par\end{centering}

\caption{Statistical results of time to extinction of cog-0 agents, for repeated
Netlogo simulations, 10,000 repeated simulations were performed for
each lattice size, the main parameters used were: maximum chromosome
length, 20; maximum plantoid reservoir capacity, 10 units; maximum
(base) m-agents' reservoir capacity, 10 units; mutation probability
0.001; replenishment rate, 2 units; replication probability, 0.9;
inital number of m-agents 20 (divided in 10 cog-0 and 10 cog-1).}
\end{table}

The skewness of the time to extinction distribution shows a positively
skewed distribution which means that the lower time to extinction
values tend to dominate, with a few less probable situations in which
the cog-0 live for longer periods. The skewness tends to become smaller
as the lattice size increases. Also, regarding the minimum time to
extinction, the values are situated bewteen 2 to 4 ticks, a tick constituting
a whole computing cycle that ends with the poison release and plantoid
replication. The maximum, on the other hand, is situated between 60
and 89 ticks.

In the same way as increasing lattice size influences positively the
cog-0 populations' resilience, so does the maximum chromosome length.
The following figure shows the boxplots, for the cog-0's time to extinction,
from 10,000 repeated simulations, for maximum chromosome lengths of
5, 10, 15 and 20, respectively. Each distribution shows evidence of
outliers to the right, of the boxplot, while the box is almost at
the center of the lowest and highest non-outlier values. The distribution
for the simulation data, obtained for maximum chromosome length of
5, shows a tendency to concentrate on lower time to extinction, which
indicates a smaller resilience for cog-0 replicators, in this case.

There is, also, a lower outlier for the length 5 data, which is a
case where, at the first simulation round, all the cog-0 became extinct.
Excluding the outliers, the boxplots still show the tendency of lower
maximum chromosome lengths to lead to the lower extinction times,
while for the other larger chromosome lengths there seems to be a
greater convergence in distribution.

\begin{figure}[H]
\begin{centering}
\includegraphics[scale=0.45]{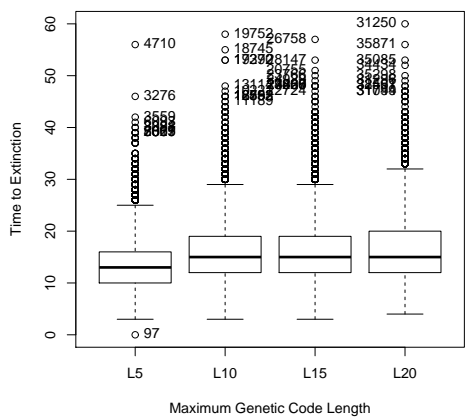}
\par\end{centering}

\caption{Boxplots of time to extinction of cog-0 agents, for repeated Netlogo
simulations, 10,000 repeated simulations were performed for maximum
chromosome lengths of 5, 10, 15 and 15 the main parameters used were:
maximum plantoid reservoir capacity, 10 units; maximum (base) m-agents'
reservoir capacity, 10 units; mutation probability 0.001; replenishment
rate, 2 units; replication probability, 0.9; inital number of m-agents
20 (divided in 10 cog-0 and 10 cog-1), lattice size 19x19.}
\end{figure}

Thus, while in harsh environments the cog-0 always go extinct, there
are factors that may contribute to a higher resilience. The critical
parameter is the plantoids' maximum reservoir capacity, if this parameter
is increased for sufficiently high values, there start to occur cases
of stabilization with sustainable coexistence of cog-0 and cog-1 agents,
however, in harsh environments the cog-1 are favored.

Group coevolutionary dynamics is a key factor in cog-1 agents' survival,
indeed, by simulating the model only with cog-0 one finds, in repeated
simulations of the model, that they always go extinct in harsh environments,
while cog-1 agents may survive and even prosper. This result does
not critically depend upon the number of letters in the formal alphabet,
thus, for two-letter, three-letter, four-letter or $n>4$-letter alphabets
we still get the same profile.

We now consider the resilience of cog-1 populations addressing a low
plantoids' maximum energy reservoir capacity corresponding to harsh
environments. While cog-0, as shown above, become extinct at this
level of harsh conditions, the cog-1 agents show two dynamical patterns:
\begin{itemize}
\item Logistic-like growth towards a high number of agents, which form small
groups and large clusters corresponding to feeding colonies;
\item Exponential growth followed by a staircase-like breakdown with a very
long time of persisting small groups.
\end{itemize}
In the first pattern, the cog-1 agents prosper, even despite the harsh
conditions, the feeding colonies seem to play a role in this, providing
energy resources that allow the agents to persist for longer times
and yielding sustainable high populations of cog-1 agents. In the
second pattern, there is an initial exponential growth but the cog-1
agents do not form large colonies, rather they peak in number and
then fall progressively in numbers to very low groups (usually below
6 individuals), these groups tend to move together and survive for
very long spans of time, they do not gather enough energy to be able
to replicate, however, they are able to gather enough energy moving
together and feeding off plantoids' and each other's energy. The survival
time spans of these m-agents surpasses largely the survival time spans
of cog-0 agents who become extinct when they reach such low numbers,
without sufficient energy to replicate.

The key factors lie, on the one hand, in the cog-1 agents' adaptive
cognition, which leads to a better energy management and, on the other
hand, in the emergent aggregation dynamics, but while the better energy
management is indeed relevant, when the cog-1 groups break up, individuals
tend to die sooner, which means that the aggregation in groups provides
a key factor in small population resilience.

While only a logic of competition was introduced, the cog-1 seem to
coevolve to make emerge aggregation and a primitive form of cooperation
not present in the individual programming, this is a collective effect.
Some groups are formed by cog-1 agents and their replicas but this
is not a condition for group formation, individuals with different
chromosomes may aggregate, groups may even merge and form larger groups
or break up, so that aggregation is not fixed.

As an illustrating example, in a simulation with maximum chromosome
length set to 20, with a two-letter alphabet, an initial number of
10 cog-1 agents, with maximum energy reservoir of plantoids chosen
set to 10 as well as m-agents' maximum base reservoir capacity set
to 10, a replicating chance of 0.9, a mutation probability of 0.001,
and a $21\times21$ torus, we found an quick growth to 20 cog-1 agents
which then started to fall, remaining 6 replicators only, which moved
together, did not replicate and continued to be alive for a very long
span, so that at 251,500 ticks two cog-1 agents died, four cog-1 agents
still remaining alive at 370,000 ticks, which still joined in groups,
also breaking up from time to time. These large survival time spans
of low density populations with low energy resources (therefore not
being able to replicate), are not found in cog-0 agents, such low
population numbers eventually tend to become extinct, however, the
time it takes for them to become extinct, in the case of cog-1 agents,
tends to largely surpass the time for extinction of cog-0 agents.

There is an emerging mutualism associated with the small cog-1 groups
that is also present in the feeding colonies' dynamics. The figure
below shows an example of a simulation only with cog-1 agents, with
a very extreme condition of 5 maximum energy units of plantoids' reservoir
capacity. In this case, of an even harsher environment than that with
10 maximum energy units, the cog-0 cannot survive, but the cog-1 can
prosper, in the figure below an example is shown, where several small
number groups coexist with feeding colonies present at several sites.
The simulation started with 20 cog-1 agents, and the population increased
to a fluctuation band between 200 and 300 m-agents. The image was
taken from the 4,000th tick.

In the figure are some clusters, where there appear more than one
m-agent), and many {}``apparent'' individual agents scattered throughout
the territory, this is apparent because most of these individual agents
are actually superimposed groups of agents that move in tandem like
one, and are only identifiable through local inspection of the patch,
this is a feature of the cog-1's nomadic groups.

\begin{figure}[H]
\includegraphics[scale=0.35]{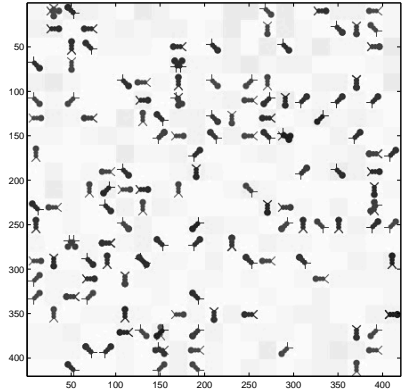}

\caption{Netlogo simulation example, figure corresponds to the 4000-th tick,
in a simulation with parameters: maximum chromosome length, 20; maximum
plantoid reservoir capacity, 5 units; maximum m-agents' (base) reservoir
capacity, 10 units; mutation probability 0.001; replenishment rate,
2.5 units; replication probability, 0.9; inital number of m-agents,
10 cog-1 agents, lattice size 21x21. The apparent dispersed individual
agents are actually groups of agents superimposed on the same patch.}
\end{figure}

For illustrative purposes, the following figure shows another simulation
with the same parameters but a maximum plantoid reservoir capacity
of 80 units. While the behavior is the same in the formation of groups
and large clusters and the nomadic groups, the population numbers
reach a band between 600 and 700.

\begin{figure}[H]
\includegraphics[scale=0.35]{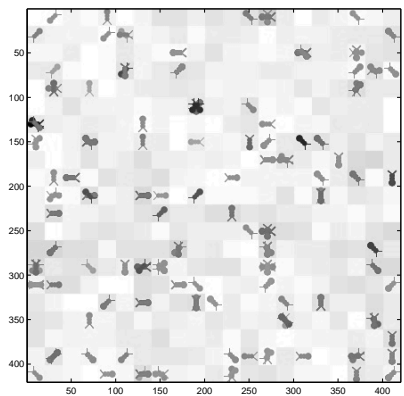}

\caption{Netlogo simulation example, figure corresponds to the 4000-th tick,
in a simulation with same parameters as figure 2, but with maximum
plantoid reservoir capacity of 80 units instead of 5.}
\end{figure}

There is a form of mutalism that emerges in the model, in the sense
that the programmed predatory procedure and the agents' motion seem
to make the group formation and large clusters a key for survival,
allowing for a group-level energy management that prevent extinction
in harsh environments, where the cog-0 agents cannot survive.

\section{Emotional responses and artificial systems}

The development of research on emotions in artificial systems can
be linked, on the one hand, to the development and introduction of
robots and intelligent systems that can interact better with us in
society, and, on the other hand, to the development of technologies
with greater adaptability to complex scenarios, the current article
explores this last venue, dealing with an artificial ecology of coevolving
artificial agents capable of basic reflexive responses that exhibit
a cognitive dynamics that can be addressed analogously in terms of
paralellism, in human organisms, to emotional responses in survival
contexts, in this case, dealing with basic survival situations: desire
to feed, fear of conflict and desire to replicate.

One cannot state that the artificial cog-1 agents of the model exhibit
emotional responses like humans, nor that they are self-aware in the
way humans are \cite{key-9}, these are different natures that we
are dealing with, different systems, however, the parallelism holds
in the sense that we introduced an artificial somatic calculatory
ability that programs the agents to respond adaptively to their internal
state and to the virtual environment in which they {}``live''.

The cog-1 agents, in the Echo-based model, have a survival-directed
cognition, namely, they evaluate their internal state and address
the environment, addressing it in a way that increases their survival
chances. Their desire to feed leads them to look for places that have
higher quantity of energy resources, however, before moving to a place
to harvest they evaluate how close they are to death, and only move
if they find that they have enough energy to do that. In predatory
situations they also anticipate the outcome of confrontation and only
attack if they find that they will benefit more from that confrontation. 

Each activity is mediated by a calculatory system that introduces
a somatic computation that leads to a level of awareness distinct
from that of the biological system but that nonetheless can be stated
in regards to the artificial system which is built in a likeness to
the biological system. We can state, for instance, that a cog-1 is
aware of its own feeding needs because it produces a cognitive synthesis
(calculated from its energy requirements) upon which it can act towards
increasing its chances of survival.

The fact that such a cognitive synthesis is produced by mathematical
rules, performed by the agent, in nothing diminishes the fact that
the agent has indeed produced a cognitive synthesis upon which it
may act, and that is supported by an organizing activity that acts
intentionally according to the rules of the system (the system's laws
of structure), aimed at the sustainable survival of the system itself,
in this sense, as argued in the main text, one can speak of a basic
form of self-awareness.

Self-awareness in artificial systems can, thus, be linked to an internal
evaluation of the system playing an adaptive role in the system's
processing and activity. In the Echo model we are dealing with programmed
response types that the agents' AI trigger automatically when faced
with adaptive problems, these response types link the agents' information
processing regarding their needs to the environmental conditions for
survival directed actions, through cognitive syntheses that are performed
through a survival directed calculatory dynamics which plays a role
akin to the role that emotions play in human organisms and in human
decision making \cite{key-7}. These can, thus, be stated to constitute
artificial emotional responses in the sense that they correspond,
in the artificial organism, to a dynamics that parallels the human
emotional responses in similar adaptive frameworks.

Thus, the organism's adaptive cognition triggers a feeding desire
response which will influence the organism's calculatory basis for
interaction with the environment (in this case, involved with moving
and feeding interactions with other agents: plantoids and other m-agents).
The generalization of this approach to artificial intelligence, in
appropriate adaptive contexts, may be useful to agilize a context-dependence
and basic survival sensible solutions on the part of artificial systems,
making them closer to the adaptive plasticity of biological systems.

Further experiments in the Echo framework for complex adaptive systems'
research and in other theoretical and applied frameworks, ranging
from ALife and multiagent simulations to robotic systems (which possess
a physical body and, therefore, a physical interface) are necessary
to evaluate the potential effectiveness in different frameworks of
such an approach.

In each of these cases Echo, general ALife, multiagent simulations
and robotics, there is a common ground: the agent is addressed as
an integrated whole and a somatic computation dynamics is introduced.
Somatic computation, which underlies artificial emotional responses,
entail the need to address the agent as an integrated whole, taking
into account the agent's body (\emph{soma}) and interface with the
environment, which includes perception and physical action. In this
sense, we are dealing with the agent in terms of an organism, and
the agent's cognition as analogous to an organism's cognition. The
Kismet robot, as well as the present Echo-based model, already integrate
some of these elements.

While, in the Echo-based model, the cog-1 agents' emotions are directly
connected to basic survival problems linked to activities necessary
to keep the agents alive, in particular, linked to feeding, moving,
interacting with other agents and replication, Kismet's emotional
responses are based upon a basic sociability problem, so that the
Kismet adapts to social interaction and sensory stimuli \cite{key-10}.

The interface and type of adaptive problems are central in what regards
emotions and somatic processing, so that a few basic questions always
need to be taken into account: what type of body does an agent possess?
What are the agent's basic adaptation problems and how do they relate
to its body's functionalities? How does the agent relate to the environment?

The type and programming of emotional responses depend upon answering
questions such as those above. Thus, different bodies imply different
interfaces with the environment, different functionalities and different
basic adaptive responses. Emotions are placed at the level of an integrated
dynamics that involves a perception system, a self-regulatory system
and an adaptive computation system.

In our case, the cog-1 creatures possess preprogrammed bodies and
{}``live'' on an artificial simulation environment, generated by
a computer, their survival problems are linked to programmed rules
of a game that is based in biology: {}``life'', in the artificial
environment, is linked to presence and existence in that environment,
death can take place in the simulation in the sense that the artificial
creature is removed from the simulation, no longer is present, no
longer exists in the simulation.

Kismet's problem is a sociability problem. Being a robot, Kismet is
a physical system, it does not reside in an artificial simulation
whose rules were programmed. Artificial simulations can be set up
so as to change the rules to differ from those of the physical world,
a robot is an existent in the physical world.

The robot's interface is a physical interface and the relation between
functionality and adaptation problems is directly built in the robot:
the robot is built with certain functionalities, perceptive systems
and computational responses to a physical environment \cite{key-10}.

Kismet was built so as to be capable of social interaction and expressing
emotional responses, mimicking recognizable human facial expressions
which become the external expressions of what, in terms of human parallelism,
constitute basic emotional responses to external stimuli%
\footnote{As illustrated in http://www.ai.mit.edu/projects/humanoid-robotics-group/kismet/kismet.html.%
}.

The basic emotional responses are associated, in Kismet, with the
robot's artificial homeostatic regulatory dynamics. The basic needs
of the robot are represented by basic {}``drives'', when the Kismet's
needs are met, the intensity level of each {}``drive'' is set within
a certain {}``desired regime'', on the other hand, as the intensity
level moves away from the homeostatic regime, Kismet's motivation
leads it to a stronger engagement in behaviors that restore that {}``drive''%
\footnote{As explained in http://www.ai.mit.edu/projects/sociable/kismet.html%
}.

Emotions, for Kismet, are based on appraisals of benefit or detriment
of a given stimulus, so that the robot evokes what are programmed
positive {}``emotive responses'' that lead it closer to the stimulus,
or negative {}``emotive responses'' that lead it away from the stimulus.
Six basic {}``emotive responses'' are modelled, as artificial analogs
of anger, disgust, fear, joy, sorrow and surprise, along with three
more arousal-based responses that correspond to interest, calm and
boredom%
\footnote{As explained in http://www.ai.mit.edu/projects/sociable/kismet.html.%
}.

The major difference between Kismet and the cog-1 agents, at this
level, is that Kismet is intended at a self-organizing communication
circuit with a human agent, that is, the {}``emotive responses''
are aimed at promoting empathy from a human caretaker, playing a central
role in regulating social interaction with the human, so that these
are aimed at future human and robot interactions.

The cog-1 agents' emotional responses are aimed at adaptive scenarios
towards survival, they are associated with decision making towards
solving adaptive problems about keeping an autonomous agent in a coevolutionary
artificial environment. Sociability was an emergent pattern in cog-1,
which is mostly engaged in its own survival, in that engagement, it
is able to spontaneously aggregate with other cog-1 agents forming
cohesive long term groups that move as if they are one larger adaptive
agent, a central behavioral pattern associated with aggregation in
complex adaptive systems \cite{key-5}.

A cog-1-like robot would be built not necessarily towards a human
and robot social context interaction like Kismet, it could be built
so that it would be furnished with a basic survival kit towards autonomous
action, including self-preservation and adaptation to changing environments.

Damásio's research has shown that, in human agents, emotions play
a central role in decision making allowing it to be faster and more
context sensitive, increasing adaptiveness in decision making problems
\cite{key-7}. Emotions, in humans, allow us to respond to threats
and identify opportunities, to decide more quickly as stressed by
Kaku \cite{key-11}. Without emotions, we would become paralized by
endless decisions all with the same weight, in this sense, artificial
agents that have a cognitive processing analogous to basic emotional
responses, incorporated in their adaptive computing systems, may prove
central for the development of robots capable of making judgments
in complex situations, including, for instance, rescue situations
linked to natural disasters.

Another application, on the other hand, to which the present work
is ultimately aimed at, is multiagent modelling: being able to incorporate
emotional responses in artificial agents may allow applications of
multiagent modelling to human systems with greater predictive ability,
which may prove useful not only to social sciences but also to industrial
applications (for instance, infrastructure design that needs to take
into account human behavior).

\end{document}